\documentclass[pdflatex,sn-nature,Numbered]{sn-jnl}

\usepackage{graphicx}%
\usepackage{multirow}%
\usepackage{amsmath,amssymb,amsfonts}%
\usepackage{amsthm}%
\usepackage{mathrsfs}%
\usepackage[title]{appendix}%
\usepackage{xcolor}%
\usepackage{textcomp}%
\usepackage{manyfoot}%
\usepackage{booktabs}%
\usepackage{algorithm}%
\usepackage{algorithmicx}%
\usepackage{algpseudocode}%
\usepackage{listings}%

\begin{document}

\title{Deep learning waterways for rural infrastructure development}
\author*[1,2]{\fnm{Matthew} \sur{Pierson}}\email{matthew.pierson@colorado.edu}

\author*[1,2,3]{\fnm{Zia} \sur{Mehrabi}}\email{zia.mehrabi@colorado.edu}

\affil[1]{\orgdiv{Better Planet Laboratory}, \orgname{University of Colorado Boulder}, \orgaddress{\city{Boulder}, \postcode{80309}, \state{Colorado}, \country{USA}}}

\affil[2]{\orgdiv{Department of Environmental Studies}, \orgname{University of Colorado Boulder}, \orgaddress{\city{Boulder}, \postcode{80309}, \state{Colorado}, \country{USA}}}

\affil[3]{\orgdiv{Mortenson Center for Global Engineering and Resilience}, \orgname{University of Colorado Boulder}, \orgaddress{\city{Boulder}, \postcode{80309}, \state{Colorado}, \country{USA}}}

\abstract{
    Surprisingly a number of Earth’s waterways remain unmapped, with a significant number in low and middle income countries. Here we build a computer vision model (WaterNet) to learn the location of waterways in the United States, based on high resolution satellite imagery and digital elevation models, and then deploy this in novel environments in the African continent. Our outputs provide detail of waterways structures hereto unmapped. When assessed against community needs requests for rural bridge building related to access to schools, health care facilities and agricultural markets, we find these newly generated waterways capture on average 93\% (country range: 88-96\%) of these requests whereas Open Street Map, and the state of the art data from TDX-Hydro, capture only  36\% (5-72\%) and 62\% (37\% - 85\%), respectively. Because these new machine learning enabled maps are built on public and operational data acquisition this approach offers promise for capturing humanitarian needs and planning for social development in places where cartographic efforts have so far failed to deliver. The improved performance in identifying community needs missed by existing data suggests significant value for rural infrastructure development and better targeting of development interventions.
}
\maketitle

\section{Main Text}
Water is fundamental to human existence, and how water flows across landscapes has important implications for human mobility \cite{BLACK2013S32,PREGNOLATO201767}.  
It is a surprising fact that not all waterways on the Earth are mapped. Part of this is due to a focus on perennial rivers and streams \cite{Messager2021-yp}, and part is due to a legacy effect of modern cartographic effort, resources, and capacity by countries to undertake mapping programs or digitize existing maps that exist. We see more comprehensive documentation in Northern American and Europe, for example, with much less understood water systems in other regions, particularly in Africa \cite{Lindersson20}.
While data intensive advances in mapping aspects like surface water have been made in recent years that allow for expanding the extent of maps of waterways, they are limited in resolution and capture only major waterways and water bodies \cite{Allen2018}; see also \cite{Pekel2016-bt,Yamazaki2019} .
A number major scientific efforts to map waterways have been undertaken and continue, including estimates of seasonal reaches \cite{Messager2021-yp}, but known gaps exist in coverage and completeness \cite{Lehner2013-sy, Yan2019-ln,Yamazaki2019}.
Gaps in knowledge of the world's waterways in data scarce areas pose issues to advancing  science for those geographies, as well as downstream tasks such as understanding impacts of climate change, infrastructure development, transport, as well as water security.

To date, waterways mapping has basically fallen into two camps – those that depend predominantly on digital elevation models (DEM) \cite{TDXHydro}, \cite{Lehner2013-sy}; and those that rely primarily on visual information from satellite imagery  \cite{Pekel2016-bt,Moortgat2022}. 
Here, we have merged these two basic ideas in a deep learning model that is scalable across large geographic areas, using publicly available satellite missions such as 10m Sentinel-2 imagery, and the 30 meter Copernicus DEM as features and  trained on labels from waterways data from the United States \cite{nhd}.
The model, WaterNet, is a computer vision based convolutional neural network built on an architecture of similar style to U-Net \cite{ronneberger2015unet}, originally developed by engineers for segmentation in biomedical research.
We show that we are able to use this model, alongside additional post-processing steps to vectorize the output, to reproduce best in class datasets of waterways not used in training \cite{TDXHydro}.
We deploy this model to map high resolution waterways wall-to-wall in a number of African countries, and are able to show that compared to existing datasets, it better addresses on the ground infrastructure needs.

We compare the outputs of this model to existing datasets. 
We note that the existing understanding of coverage of waterways by world region depends strongly on the data source.
OpenStreetMap (OSM), a dataset which is widely used, but highly dependent on community effort, demonstrates this well, with the percentage of WaterNet's inner segment points (ie all points in the polyline excluding the tip and tail, or the midpoint if there is only a tip and tail) within 0.002 degrees of an OSM waterway ranging from 43\% (Standard Deviation = 12\%) for countries in Europe, and 37\% (20\%) for basins in the United States, compared to just 18\% (12\%) for countries in Africa.
Contrast this with nations with specific mapping initiatives, such as the United States' National Hydrography Dataset (NHD) \cite{nhd}, where the gap is small at 83\% (7\%).
Notably, these geospatial differences in coverage between geographies are substantially reduced through comparisons with TDX-Hydro (TDX), a globally consistent data set from a derived from the 12m TanDEM-X DEM by United States National Geospatial-Intelligence Agency (Figure 1).

\begin{figure}
    \centering
    \includegraphics[width=\textwidth]{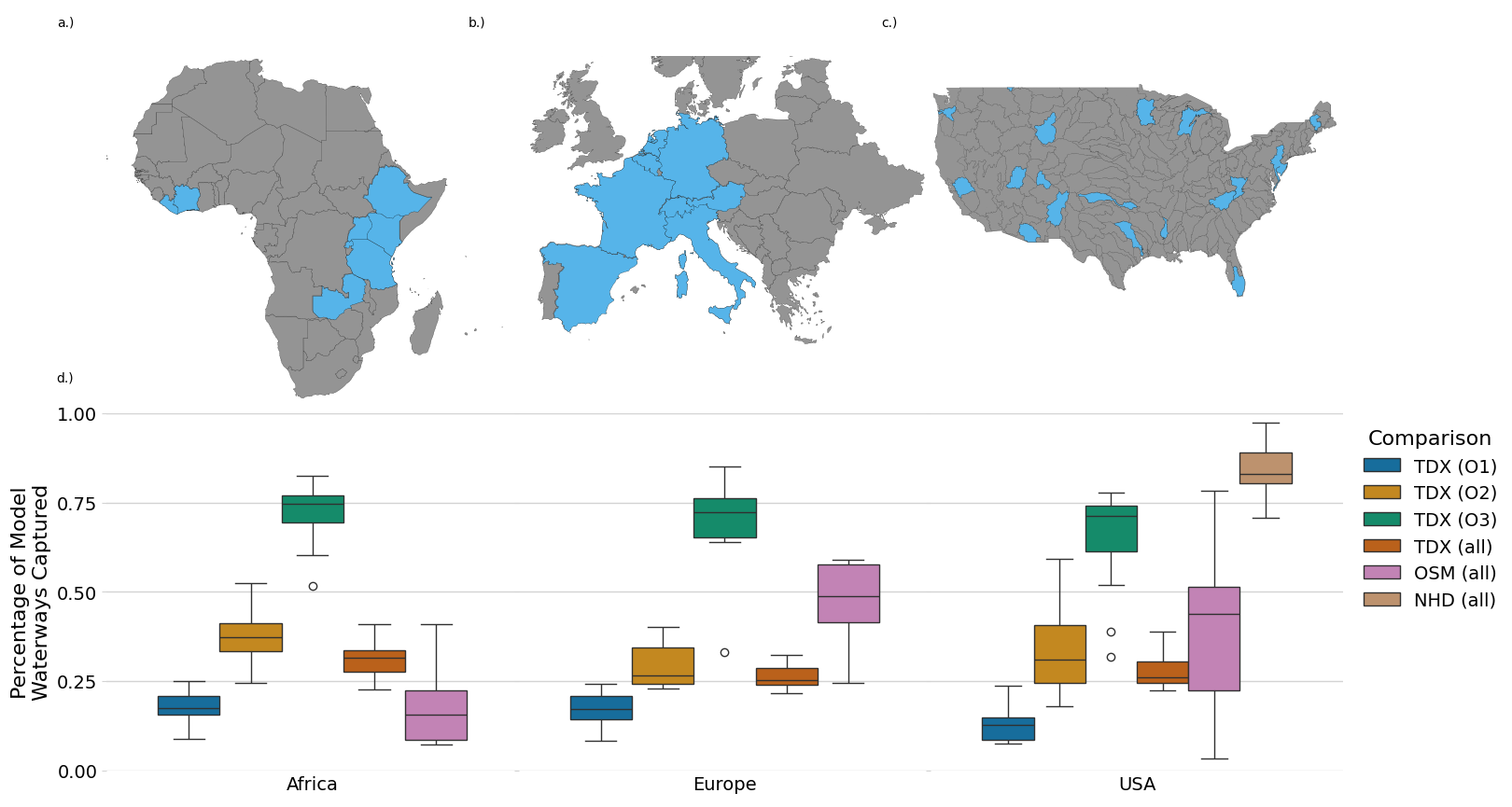}
    \caption{Potential data gaps in waterways representation in select countries in Africa. a-c) show the countries (Liberia, Côte d’Ivoire, Ethiopia, Rwanda, Uganda, Tanzania, Zambia, Kenya, Spain, France, Belgium, Netherlands, Germany, Switzerland, Austria, Italy) or watersheds used for comparisons. d) shows the correspondence between WaterNet and existing widely used, and gold standard datasets, showing substantially weaker coverage in African geographies, than countries in western Europe or watersheds in the USA, as well as lower coverage on lower stream orders. The percentage of WaterNet inner segment points at the specified stream order (O1, O2, O3, all) within 0.002 degrees of the corresponding dataset are shown.}
    \label{fig:1}
\end{figure}

Interestingly, the ability for  WaterNet to draw NHD waterways for basins where the model has not been trained, suggests potential to use computer vision to reproduce existing high quality waterways products.
An important question is whether WaterNet generated waterways can match or are able to advance on best available global products for real world use cases, outside the United States.
Importantly, we find that WaterNet is able to reproduce these high resolution streams in Africa, but adds additional detail, see visual example in Figure 2.
Detailed inspection of these maps  across the 7 African countries we have deployed in illustrate significant stretches of potential waterways that are missed not only from OSM, but also from the much more extensive TDX.
As we report below, missing these potential waterways in turn has important implications for humanitarian needs assessments in these countries.

\begin{figure}
    \centering
    \includegraphics[width=\textwidth]{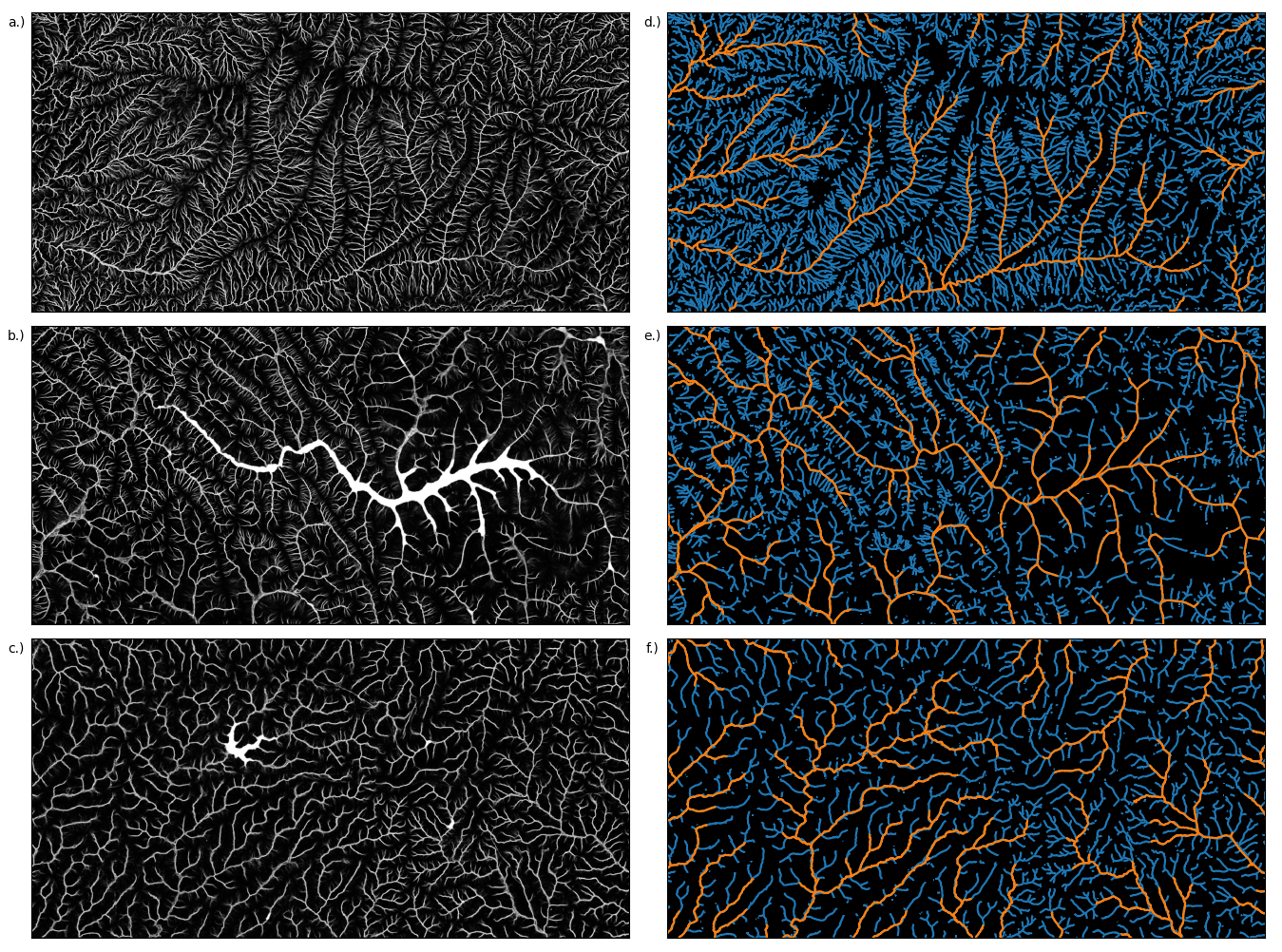}
    \caption{Example model outputs from the deep learning drawn waterways. a, b, and c are the model’s raw raster outputs with the following respective bounding boxes, Ethiopia (38.94, 11.9191, 39.46, 12.1791), Rwanda (30.04, -1.98, 30.56, -1.72) and Côte d’Ivoire (-5.4, 7.5517, -4.88, 7.8117). c, d, and f are the deep learning model’s vectorized outputs (in blue) with TDX-hydro (in orange).}
    \label{fig:2}
\end{figure}

One of the most challenging parts of mapping water in new geographies is the sparsity of publicly available ground truth data.
Here we attempt to overcome this issue, and assess the value of using WaterNet, by using independently retrieved requests for rural trail bridges.
These requests were generated through independent community consultation by the rural infrastructure organization Bridges to Prosperity, (https://bridgestoprosperity.org/)  and represent needs assessments for communities unable to access key services including education facilities, healthcare facilities, and agricultural markets due to travel obstruction posed by waterways at some point during the year. 

This validation set represents a bridging of ‘eyes in the sky’ to ‘boots on the ground’ where the validation is grounded not in scientifically determined need per se, but in local communities’ assessment of whether waterways should be mapped based on their own need for bridges to cross them. 

\begin{figure}
    \centering
    \includegraphics[width=\textwidth]{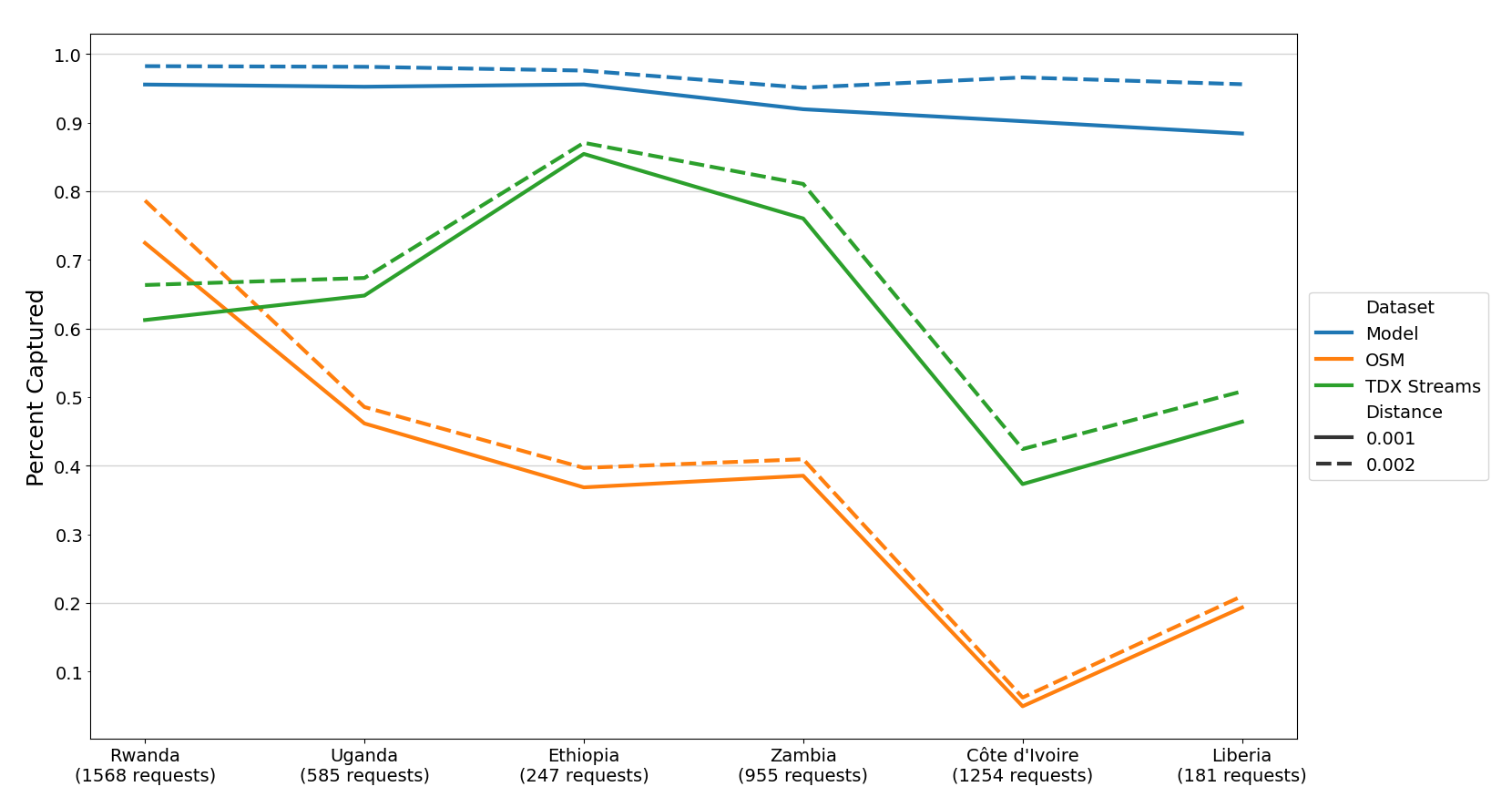}
    \caption{Community requests for infrastructure are well captured by our model. Comparisons with OpenStreetMap (OSM) and TDX Streams are given for comparison, showing the gap in community needs and what current scientific mapping exercises have been able to map.}
    \label{fig:3}
\end{figure}

Across the majority of the African countries we assess, we find the best available baseline datasets  miss a number of the local community’s infrastructure needs. More specifically, as expected, OSM performs inconsistently with a large range across countries, with 5\%-72\% of needs captured.
TDX performs better, particularly in some geographies such as Ethiopia and Zambia, were we see 76\%-85\% of requests captured, but is deficient in others e.g. Rwanda, Uganda at 61\%-65\%, and seriously so in others e.g. Liberia and Côte d’Ivoire, where more than 50\% of community requests are missed.
By way of comparison, WaterNet consistently does well and captures 88\%-96\% of the community bridge requests on the ground across countries (Figure 4).
This is critical because recall is the primary metric which we would hope these models optimize for – missing an infrastructure need where it exists carries major issues for rural inclusion in development policy around connectivity and service provision. For the countries where WaterNet performs better than baseline, it represents AI capturing of local community needs that would otherwise go overlooked by the scientific community.  

A potential criticism of these results is that they are simply overestimating waterways, as a naive model predicting waterways everywhere in the world would by definition have 100\% recall.
But detailed comparisons in places where waterways are well mapped globally shows this is not the case. For example, in the USA, comparisons with the NHD are indicative of high precision across stream orders (83\% of WaterNet's waterway inner segment points can be found within 200m of an NHD waterway, and 74\% within 100m) (Extended Data Table 1).
This indicates that our model is not overestimating where waterways should be drawn in locations where good waterways data exists.
While discrepancies in co-occurrence with NHD exist, perhaps unsurprisingly, in the lowest order stream lengths, visual inspection of the NHD data on RGB imagery at those sites also reveals missing stretches in that benchmark which WaterNet picks up (examples in Extended Data Figure 1).
Critically, we have found the full range of stream orders to be  important for local communities in terms of their requests for infrastructure to help them access resources.
For example, we find that across the 7 African countries compared, 27\% of requests fell on order 1 streams, 36\% on order 2 streams, and 38\% on order 3+ streams (Extended Data Figure \ref{fig:edf3}).
Taken together, these results are supportive of WaterNet in helping to map community needs today, without compromising false positives, in the use of these waterways data for identifying new infrastructure needs in the future.

\section{Discussion }

We show how WaterNet, a deep learning model trained in the United States can be effectively deployed in the African continent, and can reproduce the outputs of the best available data through learnt satellite and elevation features with minimal processing.
We also show how this algorithm can draw waterways of importance to local communities, streams and rivers which to date have remained unmapped in existing publicly available data.
These two factors, alongside the potential for use of these data for assisting in individual infrastructure deployments at high spatial resolution, on individual waterways stretches, offer an exciting advance to complement existing applications of use of artificial intelligence for sustainable development.

While we have presented our new model results as completely independent from existing best in class data such as TDX for comparative purposes, new work may integrate these data sets, to leverage additional benefits of the higher resolution DEM used in the backbone of TDX.
A key benefit of our fusion of satellite imagery with a digital elevation model however, permits an operational deploy to map waterways, at high resolution in time series, filling an important mapping need for dynamic assessment of ephemeral waterways structures \cite{Messager2021-yp}; and also could help better evaluate how communities are impacted by disasters in near real time.

In additional experiments we have found that tuning our model to explicitly label specific water bodies (such as swamps) in training data can allow for picking up of large scale events and humanitarian disasters related to flooding in near real time (Extended Data Figure 3). 
This coupled with new advances in deep learning forecasted weather \cite{Lam2023} or flooding events \cite{Nearing2024}, provides promise for more adaptation focused infrastructure planning that maintains physical connectivity for rural communities in face of climate change and other environmentally driven disruptions to connectivity.  

Importantly, waterways mapping efforts focused on physical hydrology alone, can intentionally disregard reaches (rivers, streams etc) because the hydrological flow models themselves so far developed are not designed to represent the full network at high resolution \cite{hales}.  Integrating a social perspective on community need for mapping those reaches, might enable a better understanding of resilience, such as how extreme weather events and flooding, may cause communities to be cut off some essential resources. We show incorporating community consultation can capture those community needs more comprehensively, and as a result could potentially provide a firm foundation to develop programs that reduce inequity and exclusion of those communities from development planning around which interventions are layered. 

Recently, there has been a rise in the use of artificial intelligence for sustainable development, enabled through advancements in earth observation with high repeat frequency and high resolution sensors \cite{Burke2021-av}. Notable successful examples have been in identifying  poverty and causally evaluating the impacts of infrastructure, or public safety net programs on  income proxies {\cite{Ratledge2022-up}; \cite{Aiken2022-tl}}. However, there is a recognized need for investment and improvement of the application of these technologies by governments and the international monetary organizations, at all stages of needs assessment, implementation, and evaluation, due to their geographic reach and cost efficiency \cite{WDR2021}. 

Using computer vision to learn waterways from elevation and satellite input features conjointly also offers a new opportunity for cost effective generation of waterways at scale, one that to date has not been fully utilized. More broadly however, the work we present suggests an important opportunity for modelers and those focusing on social impact of artificially intelligence work more closely together to map biophysical structures and the interaction between humans and natural systems. We see promise for this collaboration so greater benefits of advancements in artificial intelligence and remote sensing could be used to reap more inclusive development policy in geographies where it is needed most.

\section{Methods}
\subsection{Satellite data}
Sentinel-2 (NRGB bands) \cite{sentinel2} and Copernicus DEM GLO-30 \cite{copdem} data were acquired from the Microsoft Planetary Computer API \cite{mpc}. Cloud-free composites were assembled by iteratively downloading Sentinel-2 tiles for the year 2023 and choosing dates with minimal cloud cover, as determined by Sentinel-2's scene classification. In cases where more than 1\% of the tiles pixels had clouds, cloud shadows, or missing data, additional tiles were downloaded, prioritizing those that minimized the coverage of cloud classified pixels in the final composite.  The cloud free pixels of the individual images where normalized by channel, and transformed using $$f(x) = \text{round}\left(\dfrac{255}{1+e^{-0.6x}}\right).$$ Finally, to further remove clouds unidentified by the scene classification in the transformed images, a buffer of 500m around clouds, missing data or cloud shadows were also removed, prior to mean compositing, and the images were stored as 8-bit unsigned integers. These data were used to create input features included 10 channels: the first four being transformed Sentinel-2 NRGB channels ($NRGB_t$), and the remaining 7 being $NDVI$, $NDWI$, Shifted Elevation ($E_S$), Elevation x-delta ($\Delta_x E$), Elevation y-delta ($\Delta_y E$), elevation gradient ($\nabla E$).

\subsection{Waterways labels}
The National Hydrography High Resolution Dataset (NHD) \cite{nhd} was utilized as training data. NHD is mapped at a scale of 1:24,000 scale or better in the contiguous United States (CONUS) and represents the most up-to-date and geographically inclusive hydrography dataset available for the United States, covering all basins, a wide range of terrain types and landscapes, hydrographic contexts, and biomes (Tropical Rain-forest, Tropical Deciduous Forest, Temperate Deciduous Forest, Temperate Coniferous Forest, Boreal Forest, Tundra, Temperate Grassland, Desert) – making it an interesting candidate for training a model for out-of-sample prediction in other geographies in the world outside the United States. NHDPlus encodes a wide range of features such as naturally occurring and man-made surface water (lakes, ponds, and reservoirs), water flow paths (canals, ditches, streams, and rivers), and point features (springs, wells, stream gauges, and dams). The surface water and water flow paths were  burned to rasters, with each fcode type (e.g. the fcode in the NHD data is an identifier for each water type, such as rivers, streams, lakes, ditches, intermittent, ephemeral versions of each, etc.) assigned a different integer label, which was used to give different water(way) types different weights during training.

\subsection{Model Architecture}
We build a novel deep learning model architecture following a U-Net style approach \cite{Ronneberger2015}, a computer vision algorithm original developed in the biomedical segmentation tasks. It consisted of encoder and decoder layers, with the rows and columns halved and the channels doubled for five iterations (encoder layers), followed by the reverse for three iterations (decoder layers). The initial layer included a global attention type layer, instance normalization, and a convolutional block to increase the number of channels. Encoding layers comprised 2x2 convolutions without padding, instance normalization layers, residual convolutional layers, a multiplication layer, and a fully connected layer. Decoding layers included transposed convolutions, instance normalization layers, residual blocks, multiplication blocks, a fully connected layer, and a convolutional layers.  This version of WaterNet creates raster outputs at 40m resolution.

\subsection{Model Training}
The model was trained on waterways labels using two grid sizes. First the model was trained on ~1.5M (1,543,120) 224x224 size grids with 2.4K validation grids. 
In this case, the iterations consisted of 2K training grids with 2 randomly chosen augmentations for each grid, resulting in 6k training grids. 
The augmentations consisted of reflecting and rotating the original images, and dropping out 20\% of the input cells. We used a batch size scheduler with an initial batch size of 100, which increased by 100 if the F1-score on the validation data did not increase for 15 iterations.  The optimizer was stochastic gradient descent with momentum and L2 regularization. (learning rate=0.125, momentum=0.9, weight decay=0.0001).  The model was then trained on ~96K (96,691) 832x832 size grids with 200 validation grids and no augmentations.  In this case, the iterations consisted of 400 training grids, with an initial batch size of 8, which increased by 8 if the F1-score on the validation data did not increase for 15 iterations; with the same optimizer and settings.  The loss function was Binary Cross-Entropy (BCE), weighted by fcode type (Extended Data Table \ref{tab:edt2}). Weighting allowed us to adjust for the label imbalance and also to down weight/ mask out classes that create different kinds of features in the output.  Experiments showed re weighting swamp fcode values, for example, allowed for tuning model versions to potentially pick up humanitarian disasters and flooding events (Extended Data Figure \ref{fig:edf2}).

\subsection{Post-Processing}
For evaluation the 40m raster output required vectorization. Within a given watershed or country used for deploy tiles were merged, waterways thinned and vectorized. Each cell in the raster was assigned weights based on elevation. Cells were then iteratively labeled as three states,  either  $S$ (skeleton),  $I$ (interior), or  $R$ (removable) (a point that can be removed without altering the topology) in a process of thinning. A cell is labeled $S$ cell if it is touching a max of one other waterway cell, or if its removal would change the connectedness of the waterway (if its removal would turn a single waterway into two or more waterways which were no longer connected), labeled as $I$ if its removal would introduce a gap in the waterway, and the remainder of cells labeled $R$. $R$ cells were iteratively checked and removed without altering topology, in order of elevation (highest to lowest) with neighboring interior points reassessed for label status changes at each iteration. The process completes when there are no $I$ or $R$ points remaining. Finally, midpoints of $S$ cells from the thinning process were used as points in a polyline to represent waterways.  

A modified Strahler stream order is then determined for the vectorized output, to better evaluate model performance.  Rather than starting at a leaf node, we traverse through the waterways in order by elevation, with the same rule that if two streams of the same order $n$ merge, then the result is a stream or order $n+1$, and if streams of order $n$ and $m$ merge then the resulting stream is of order $\max(n, m)$, where $n$ and $m$ equal the order of the respective connecting streams. These modifications make it possible to put an order on a non-tree graph. We also set any stream with both endpoints connected to another waterway to have a stream order of $\max(2, \text{current stream order})$. We generally use 3 categories, order 1 streams, order 2 streams, and order 3 and above streams. This is simply for comparison purposes for this study (and not intended for use of water flow routing). 

\subsection{Model deployment}
We deployed this model across all contiguous US HU4 watersheds (including 18 held completely out in model training), 8 countries in western Europe (Spain, France, Belgium, Netherlands, Germany, Switzerland, Austria, Italy), and 8 countries in the African continent (Liberia, Côte d’Ivoire, Ethiopia, Rwanda, Uganda, Tanzania, Zambia, Kenya). After post-processing, we then used these deploys to evaluate the performance of the model. 

\subsection{Evaluation}
Notably in the USA, our comparisons were direct to the input labels, being derived from the NHD, and represent standard out of sample tests for hold out basins.
In Europe and in Africa we relied on two independent existing waterways datasets to evaluate model performance. For this purpose we utilized a widely accessible data set, OpenStreetMap, and the most extensive waterways data set currently available, TDX-Hydro, developed by the National Geospatial-Intelligence Agency using the 12m TanDEM-X DEM.
Critically, we also evaluate the comparisons for OSM and TDX in the USA as a way of triangulating among these different source datasets.  
Points were taken from the model’s vectorized output excluding the tip and tail of reaches (to avoid potentially double counting those points).
We found the nearest waterway from the test datasets  (OSM, TDX, or NHD), and found the distance to that waterway in each case.
We consider hits if that distance was less than 0.001/ 0.002 degrees ($\sim$100m/200m).
We did this for stream order 1, 2, and 3 plus waterways, and all waterways.
Together this allowed us to evaluate how well the model performed in the USA with the hold out set, but also compare the model performance in new geographies against available benchmark datasets.

\subsection{Community Requests}
We independently assessed the model performance with community requests for bridge sites collected by the non-governmental organization Bridges to Prosperity (B2P) across 6 countries in Africa (Rwanda, Uganda, Ethiopia, Liberia, Côte d'Ivoire, Zambia).
In total these represent 4790 unique requests that communities have made for a bridge to enable those communities to access essential services such as schools, healthcare facilities and markets.
The spatial coverage and sampling of of these data differ by country, and range from full-coverage nationwide needs assessments (Rwanda), full coverage assessments within specific regions (Uganda, Ethiopia), full coverage within 5km buffer zones sampled within the country (Côte d'Ivoire), and opportunistic (Zambia, Liberia, Ethiopia).
Differences also exist in the order of stream coverage, although overall 25\% fell on first order streams, 35\% on second order, and 30\% on third order or greater. As our main focus was recall, differences in sampling intensity were less of a concern, although we provide country level breakdowns of performance for clarity. We assessed recall for our new computer vision generated waterways data against the benchmarks of OSM and TDX for all of the aforementioned countries.

\section{Data Availability}
Data for reproducing the results in this manuscript can be retrieved from the Harvard Dataverse repository doi:10.7910/DVN/A44GMQ, except the bridge request data which requires an independent request to Bridges to Prosperity.  We retrieved all Sentinel-2 and Copernicus DEM data from the Microsoft Planetary Computer API. Other waterways data sources used in this paper are accessible from the publicly accessible links provided.

\section{Code Availability}
Code for reproducing the model validation and inter-comparison results in this manuscript can be retrieved from the Harvard Dataverse repository doi:10.7910/DVN/A44GMQ. Publication of WaterNet model code will be available in a following publication.

\bibliography{bibliography}

\section{Acknowledgments}
The authors would like to thank Bridges to Prosperity for sharing data and offering feedback (Abbie Noriega, Kyle Shirley, Cameron Kruse, Etienne Mutebutsi, Angella Nakkungu, and field assistants).  This project was funded by Bridges to Prosperity under the grant “Remote Impact Assessment of Rural Infrastructure Development” to the Better Planet Laboratory (https://betterplanetlab.com/).

\section{Author Contributions}
MP and ZM conceptualized and designed the project. MP designed and implemented the waterways model and conducted the downstream analysis with guidance from ZM. MP and ZM interpreted the results and wrote the paper.

\section{Competing Interests}
None

\newpage
\clearpage
\renewcommand\thefigure{\arabic{figure}}    
\setcounter{figure}{0}
\begin{figure}
    \centering
    \includegraphics[width=\textwidth]{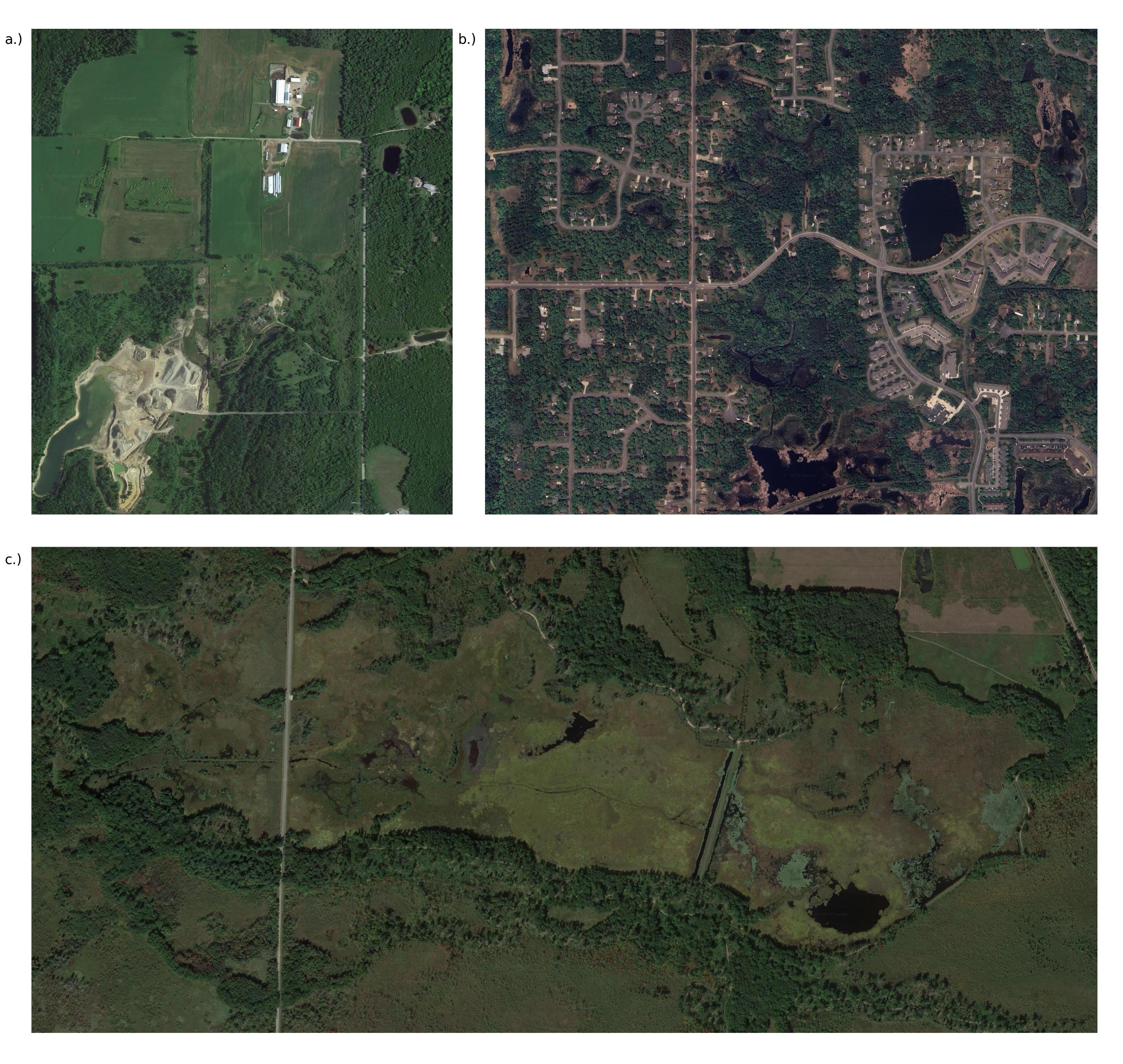}
    \def\figurename{Extended Data Figure}

    \caption{Illustrative examples of missing waterways in NHD data. The bounding boxes for a, b, and c are (-87.8104, 45.1841, -87.7965, 45.1955), (-94.2725, 46.3569, -94.2522, 46.3680), and (-88.5530, 44.6436, -88.5177, 44.6550) respectively. None of the water in these bounding boxes were labeled in the NHDPlus data.}
    \label{fig:edf1}
\end{figure}

\newpage
\clearpage

\begin{figure}
    \centering
    \includegraphics[width=\textwidth]{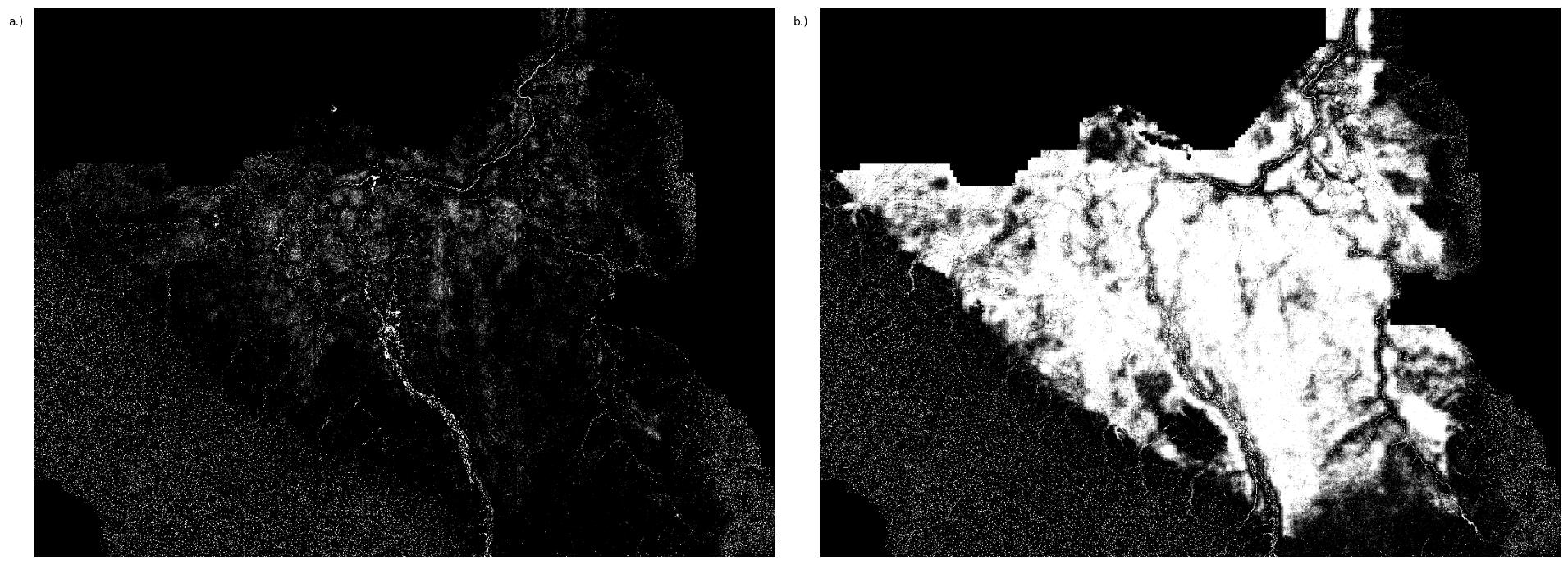}
    \def\figurename{Extended Data Figure}

    \caption{Fine tuning model for humanitarian use-case. Major 2019 South Sudan flooding leading to human displacement. Identification of event with (left) and baseline (right) adjusting fcode weights for swamps.}
    \label{fig:edf2}
\end{figure}

\newpage
\clearpage

\begin{figure}
    \centering
    \includegraphics[width=\textwidth]{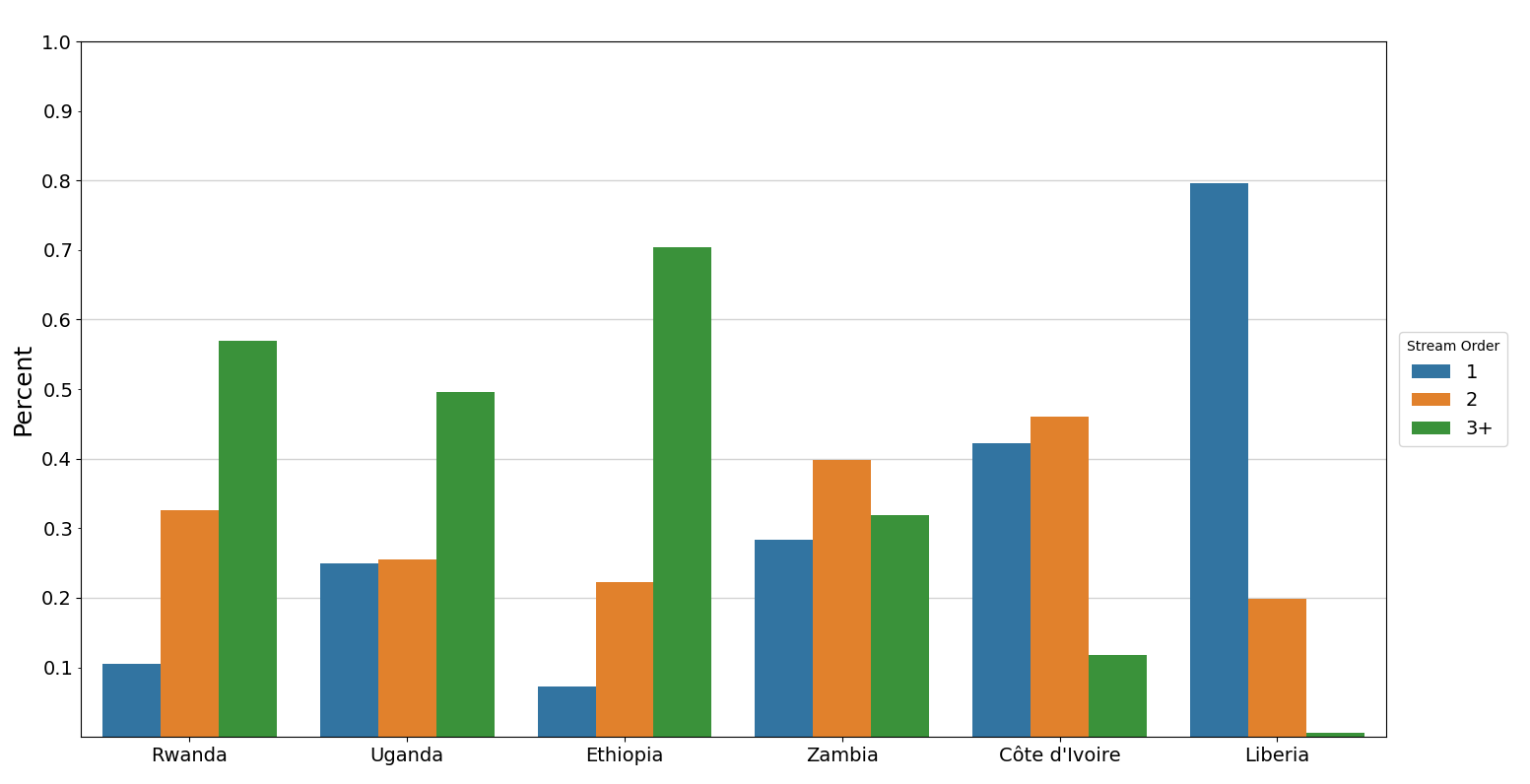}
    \def\figurename{Extended Data Figure}

    \caption{Independent community bridge site requests by country and stream order.
}
    \label{fig:edf3}
\end{figure}

\newpage
\clearpage

\begin{table}
    \centering
    \begin{tabular}{|c||c|c|c|c|}
        \hline
        Stream Order & 1 & 2 & 3 & all \\
        \hline
        Count & 12,238,517 & 8,515,920 & 3,186,058 & 23,940,495 \\
        \hline
        Mean & 145.64 & 59.53 & 34.0 & 100.15 \\
        \hline
        SD & 237.38 & 137.64 & 78.52 & 196.45 \\
        \hline
        Min & 0.0 & 0.0 & 0.0 & 0.0 \\
        \hline
        5\% & 0.0 & 0.0 & 0.0 & 0.0 \\
        \hline
        10\% & 1.45 & 0.0 & 0.06 & 0.0 \\
        \hline
        15\% & 4.41 & 0.43 & 1.77 & 2.32 \\
        \hline
        20\% & 7.47 & 2.54 & 3.51 & 4.74 \\
        \hline
        25\% & 10.7 & 4.69 & 5.29 & 7.24 \\
        \hline
        30\% & 14.2 & 6.9 & 7.11 & 9.85 \\
        \hline
        35\% & 18.12 & 9.21 & 9.0 & 12.63 \\
        \hline
        40\% & 22.75 & 11.63 & 10.97 & 15.64 \\
        \hline
        45\% & 28.67 & 14.22 & 13.05 & 19.0 \\
        \hline
        50\% & 37.13 & 17.04 & 15.28 & 22.89 \\
        \hline
        55\% & 51.39 & 20.17 & 17.68 & 27.65 \\
        \hline
        60\% & 78.06 & 23.79 & 20.34 & 33.93 \\
        \hline
        65\% & 115.31 & 28.14 & 23.35 & 43.26 \\
        \hline
        70\% & 156.81 & 33.7 & 26.88 & 60.34 \\
        \hline
        75\% & 202.77 & 41.54 & 31.23 & 95.59 \\
        \hline
        80\% & 256.2 & 54.72 & 37.0 & 150.14 \\
        \hline
        85\% & 322.85 & 84.48 & 45.65 & 218.56 \\
        \hline
        90\% & 415.33 & 158.53 & 62.76 & 311.18 \\
        \hline
        95\% & 582.65 & 304.61 & 125.44 & 468.79 \\
        \hline
        99\% & 1,060.14 & 676.35 & 390.75 & 899.76 \\
        \hline
        Max & 12,481.25 & 5,538.55 & 2,811.67 & 12,481.25 \\
        \hline
    \end{tabular}
    \def\tablename{Extended Data Table}
    \caption{Vigintiles of the distance from the inner segment points of WaterNet's waterways to the nearest NHD labeled waterway. The distance, in meters, was found from those points to the nearest waterway in the NHD dataset. 
}
    \label{tab:edt1}
\end{table}

\newpage
\clearpage

\begin{table}[]
    \centering
    \begin{tabular}{|c|c|}
        \hline
        Water(way) Type & Weight \\
        \hline
        playa & 0.0 \\
        \hline
        Inundation area & 0.0 \\
        \hline
        Swamp Intermittent & 0.5 \\
        \hline
        Swamp Perennial & 0.5 \\
        \hline
        Swamp & 0.5 \\
        \hline
        Reservoir & 0.5 \\
        \hline
        Lake Intermittent & 0.5 \\
        \hline
        Lake Perennial & 3.25 \\
        \hline
        Lake & 3.25 \\
        \hline
        spillway & 0.0 \\
        \hline
        drainage & 0.5 \\
        \hline
        wash & 1.5 \\
        \hline
        canal storm & 0.5 \\
        \hline
        canal aqua & 0.5 \\
        \hline
        canal & 0.5 \\
        \hline
        artificial path & 2.5 \\
        \hline
        Ephemeral Streams & 3.5 \\
        \hline
        Intermittent Streams & 3.75 \\
        \hline
        Perennial Streams & 3.25 \\
        \hline
        Streams Other & 3.25 \\
        \hline
        other & 0.5 \\
        \hline
    \end{tabular}
    \def\tablename{Extended Data Table}
    \caption{Model weights for fcode labels. A weight of 0 indicates the NHD data were considered to not be waterways, a weight between 0 and 1 were masked out, weights greater than or equal to 1 were used to scale the BCE loss contribution of that pixel by that amount.
}
    \label{tab:edt2}
\end{table}

\end{document}